\documentclass[lettersize,journal]{IEEEtran}
\usepackage{amsmath,amsfonts}
\usepackage{algorithmic}
\usepackage{algorithm}
\usepackage{array}
\usepackage{textcomp}
\usepackage{stfloats}
\usepackage{url}
\usepackage{verbatim}
\usepackage{graphicx}
\usepackage{subcaption}
\usepackage{cite}
\usepackage{tabularx}
\usepackage{multirow}
\usepackage{graphicx}
\usepackage{tikz}
\usetikzlibrary{shapes.geometric, arrows.meta, positioning}

\tikzset{
    block/.style={rectangle, draw, text centered, minimum height=1em, minimum width=4em, font=\small},
    arrow/.style={-Stealth, thick},
}

\begin{document}

\title{Development and Validation of a Modular Sensor-Based System for Gait Analysis and Control in Lower-Limb Exoskeletons}

\author{Giorgos Marinou, Ibrahima Kourouma, Katja Mombaur, \IEEEmembership{Member, IEEE}% <-this % stops a space
\thanks{This work was supported by the Carl Zeiss Foundation through the project HeiAge. (\it{Corresponding author: Giorgos Marinou.)}}% <-this % stops a space
\thanks{This work involved human subjects or animals in its research. Approval of all ethical and experimental procedures and protocols was granted by the Ethics Committee of Heidelberg University under Application No. S-313/2020, and performed in line with the Declaration of Helsinki.}% <-this % stops a space
\thanks{Giorgos Marinou and Ibrahima Kourouma are with the Institute of Computer Engineering (ZITI), Heidelberg University, 69120 Heidelberg, Germany (e-mail: giorgos.marinou@ziti.uni-heidelberg.de).}
\thanks{Katja Mombaur is with the Institute for Anthropomatics and Robotics, Karlsruhe Institute of Technology, 76131 Karlsruhe, Germany, and also with the Department of Systems Design Engineering and the Department of Mechanical and Mechatronics Engineering, University of Waterloo, Waterloo, ON N2L 3G1, Canada (e-mail: katja.mombaur@kit.edu).}}



\maketitle

\begin{abstract}
With rapid advancements in exoskeleton hardware technologies, successful assessment and accurate control remain challenging. This study introduces a modular sensor-based system to enhance biomechanical evaluation and control in lower-limb exoskeletons, utilizing advanced sensor technologies and fuzzy logic. We aim to surpass the limitations of current biomechanical evaluation methods confined to laboratories and to address the high costs and complexity of exoskeleton control systems. The system integrates inertial measurement units, force-sensitive resistors, and load cells into instrumented crutches and 3D-printed insoles. These components function both independently and collectively to capture comprehensive biomechanical data, including the anteroposterior center of pressure and crutch ground reaction forces. This data is processed through a central unit using fuzzy logic algorithms for real-time gait phase estimation and exoskeleton control. Validation experiments with three participants, benchmarked against gold-standard motion capture and force plate technologies, demonstrate our system's capability for reliable gait phase detection and precise biomechanical measurements. By offering our designs open-source and integrating cost-effective technologies, this study advances wearable robotics and promotes broader innovation and adoption in exoskeleton research.
\end{abstract}


\section{Introduction}

\IEEEPARstart{W}{earable} robots, particularly lower-limb exoskeletons (LLEs), have emerged as revolutionary tools in recent years for enhancing mobility in impaired individuals. These devices are critical in managing a broad spectrum of musculoskeletal disorders and injuries, significantly improving the quality of life for affected individuals \cite{duPlessis2021AAssistance, Rehmat2018UpperReview, Wang2022ADisabled}. In the vast majority of cases, arm crutches are required in order to operate the exoskeleton \cite{Smith2020EstimatingCrutches}, to provide static and dynamic stability, through transitions of sitting and standing, as well as walking. LLEs often rely on adjunct technologies, incorporating advanced sensors and control algorithms to closely mimic natural gait patterns. Despite groundbreaking advancements in sensor technology and biomechanical evaluation, the adoption of LLEs faces significant barriers. These challenges include the high costs associated with precise motion capture systems for biomechanical assessment and the complexity of integrating varied technologies into cohesive systems capable of accurately interpreting user intentions in real-time \cite{Marinou2022TowardsFamiliarization, Li2021ReviewExoskeletons}.

Current methodologies, such as marker-based motion analysis and force plate technologies, though accurate, are predominantly confined to laboratory settings due to their prohibitive costs and extensive setup requirements \cite{Barbareschi2015StaticallyHuman, Fineberg2013VerticalParaplegia}. Recently, marker-less motion capture has begun to address these limitations, offering promising results without the cumbersome setup of traditional systems \cite{Kanko2021ConcurrentCapture}. However, these technologies still struggle to operate effectively in everyday environments, a critical requirement for practical LLE applications, while still needing camera equipment to cover a certain volume.

\begin{figure}[!b]
    \centering
    \includegraphics[width=1\linewidth]{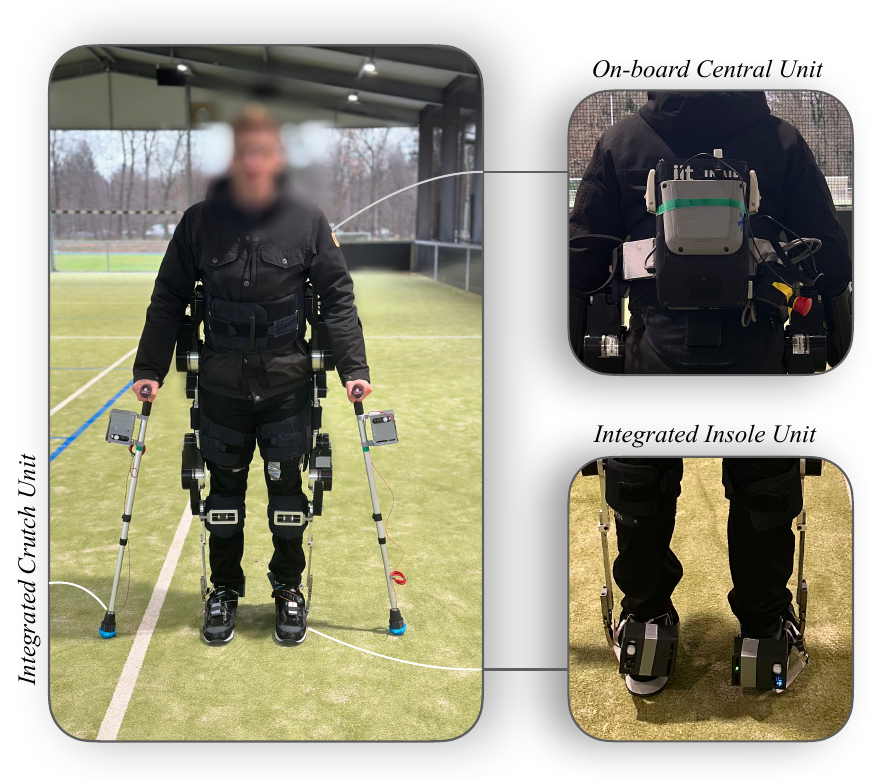} 
    \caption{Outdoors experimental setup using crutches and insoles system. The two sub-systems are integrated seamlessly with the exoskeleton while the central unit is securely mounted on the on-board computer interface of the exoskeleton in the back.
    }
    \label{fig_experiment}
\end{figure}

The integration of inertial measurement units (IMUs), despite enabling kinematic measurements and gait phase identification outside of lab settings, still faces challenges in calibration and comfort when combined with lower-limb exoskeletons (LLEs) \cite{Ribeiro2017InertialAnalysis, Park2021ValiditySPM, Seel2014IMU-BasedAnalysis, DeMiguel-Fernandez2023InertialStudy}. Similarly, surface electromyography (sEMG) quantifies muscular effort but is hindered by preparation requirements and sensitivity to environmental conditions \cite{Park2017PerformanceSignals, Kong2022ErgonomicSystem, Abdoli-Eramaki2012TheSpectrum}. Despite these limitations, both IMU and sEMG technologies serve as valuable control input when processed correctly, although unreliability issues due to environmental noise have led to exploring alternative input methods \cite{Sun2022FromReview, Netukova2022LowerState-of-the-Art, Moon2019IntentionExoskeleton, Kawamoto2003PowerController, Guizzo2005TheExoskeletons, Kawamoto2003PowerInformation, Iqbal2021NeuromechanicalRecognition, Zhang2022DesignInformation}.

\begin{figure*}[!t]
    \centering
    \begin{tabular}{cc}
      \includegraphics[width=0.3235\linewidth]{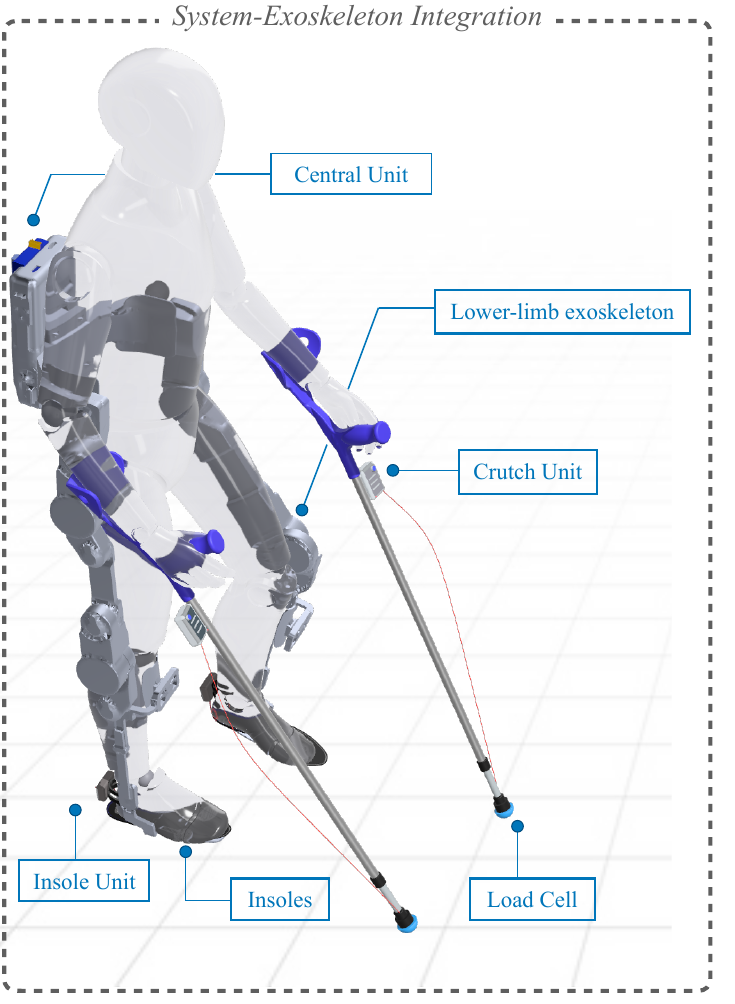} & \includegraphics[width=0.45\linewidth]{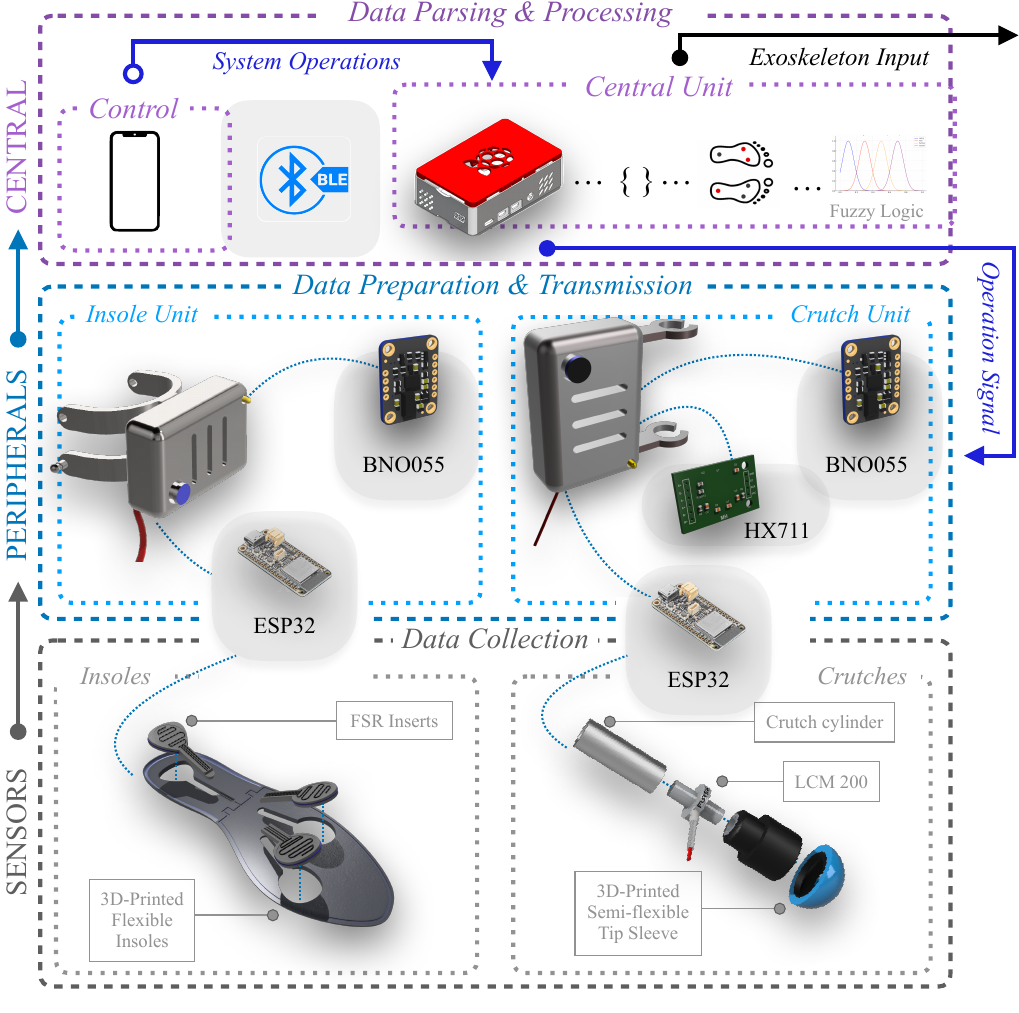}  \\
      (a) & (b)
    \end{tabular}
    \caption{Overview of System Integration and Design. 
    a) The system components integrated onto a lower limb exoskeleton include: (1) two flexible 3D-printed insoles, (2) insole data collection units, (3) forearm crutches with load cells, (4) crutch data collection units, and (5) a central data and control unit mounted on the exoskeleton’s back. b) Simultaneous data collection is achieved through a sensor framework. Modular insoles with removable FSR inserts measure pressure at critical points: the heel, first, and fifth metatarsals. Forces are captured by load cells at the crutch tips, encased in semi-flexible sleeves. Data from the sensors, processed via ESP32-S3 Feather boards, are transmitted to a central unit using BLE. This unit, powered by a Raspberry Pi 4, parses data into a CSV file, while a fuzzy logic algorithm computes gait phases. An Android application manages communication between the central unit and peripherals, overseeing data flow and system operations.
    }
    \label{fig_systemoverview}
\end{figure*}

Force-measuring crutches and pressure-sensing insoles equipped with strain gauges and piezoelectric sensors within force-sensitive resistors (FSRs) offer another way to analyze movement effectively \cite{Gil2021DevelopmentSclerosis, Khandakar2022DesignTemperature}. Integrating these with IMUs creates a versatile system that enhances user intention assessment across various platforms. Although these components are proven to effectively analyze gait in multiple studies \cite{Gil2021DevelopmentSclerosis, Khandakar2022DesignTemperature, Manupibul2014DesignActivities, Nascimento2022DevelopmentEvaluation, Gonzalez2015AnInsoles, Silva2017BuildingCities}, there is a notable gap in research combining these technologies into a unified, multi-sensor system. Furthermore, the reproducibility of these methods and hardware is often unclear and complicated, and the cost is relatively high, while lacking open-sourced protocols that hinder further research development.

To bridge this gap, our research introduces a cost-effective, easy-to-reproduce, and universally compatible sensor system designed explicitly for lower-limb exoskeletons. Unlike conventional high-cost systems whose precision exceeds practical necessities and often lacks integration capabilities for direct exoskeleton control, our solution prioritizes functional accuracy and ease of integration \cite{Yin2019PersonalisedInference, Armannsdottir2020AssessingStudy}. Our approach harnesses the power of IMUs and force sensing technologies, combined in a modular, adjustable framework that supports real-world application across various terrains \cite{Ribeiro2017InertialAnalysis, Park2017PerformanceSignals}.

We have developed a standalone and modular two-part system consisting of crutches equipped with force sensors and insoles that incorporate pressure sensors, both integrated with IMUs. This multi-sensor assembly not only captures detailed biomechanical data but also interprets user intent with high reliability, establishing intuitive control of the exoskeletons. To underscore our commitment to accessibility and community-driven improvement, we provide this technology as an open-source platform, ensuring that other researchers can adapt and enhance our system without duplicate development efforts. Further, the system can be used independently and fitted to any lower-limb exoskeleton or exosuit. In this study, we test our ssytem using the TWIN LLE \cite{Laffranchi2021}.

Moreover, we employ fuzzy logic to process sensor inputs, offering a robust alternative to traditional machine learning methods that require extensive data and often yield unpredictable outcomes. Unlike machine learning, fuzzy logic is computationally inexpensive and follows a rule-based approach, eliminating the need for large datasets that may not exist and avoiding the generation of incorrect data or control inputs for the exoskeleton. This decision-making framework, built on predefined rules, effectively handles the inherent variability in human gait, thus optimizing the exoskeleton’s response to user movements \cite{Kong2008SmoothPatterns, Zadeh1965FuzzySets}.

Building upon this premise, this study delivers a modular, mobile, and cost-effective sensor-based system for biomechanical evaluation and control of lower-limb exoskeletons. Our standalone system is designed for easy integration with any exoskeleton, enhancing real-world applicability by enabling evaluations outside laboratory settings. We also provide full access to our software and hardware designs to promote collaboration and innovation in the field. In summary, this study not only addresses critical challenges in motion intention detection and the integration of sensor technologies into lower-limb exoskeletons in a low-cost manner but also sets a new standard for open-source contributions in wearable robotics. Our approach significantly lowers the barriers to entry for researchers and developers, fostering broader innovation and practical applications in the field.

The following sections detail the design and methods used in our system, providing a comprehensive guide for replication and adaptation included in the supplementary materials, alongside validation experiments that benchmark our technology against current standards. The remaining sections are structure as follows: Section II introduces the standalone system and the design of its independent components, Section III describes the validation experiments, Section IV reports the results of these experiments and discusses their significance, Sections V outlines the system's potential applications with LLEs, and Section VI concludes the current study. 

\begin{figure*}[!t]
    \centering
    \begin{tabular}{ccc}
      \includegraphics[width=0.45\linewidth]{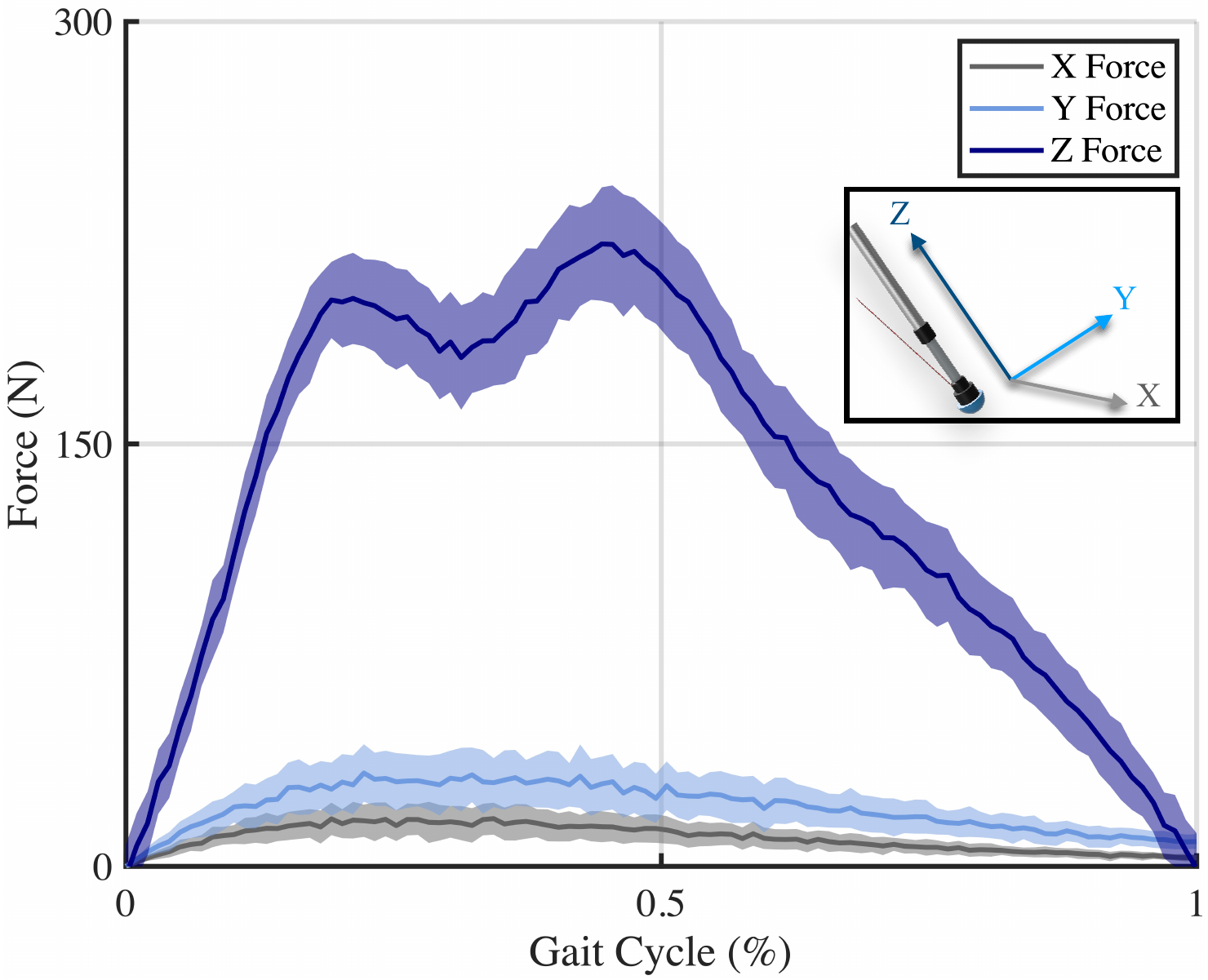} & \includegraphics[width=0.485\linewidth]{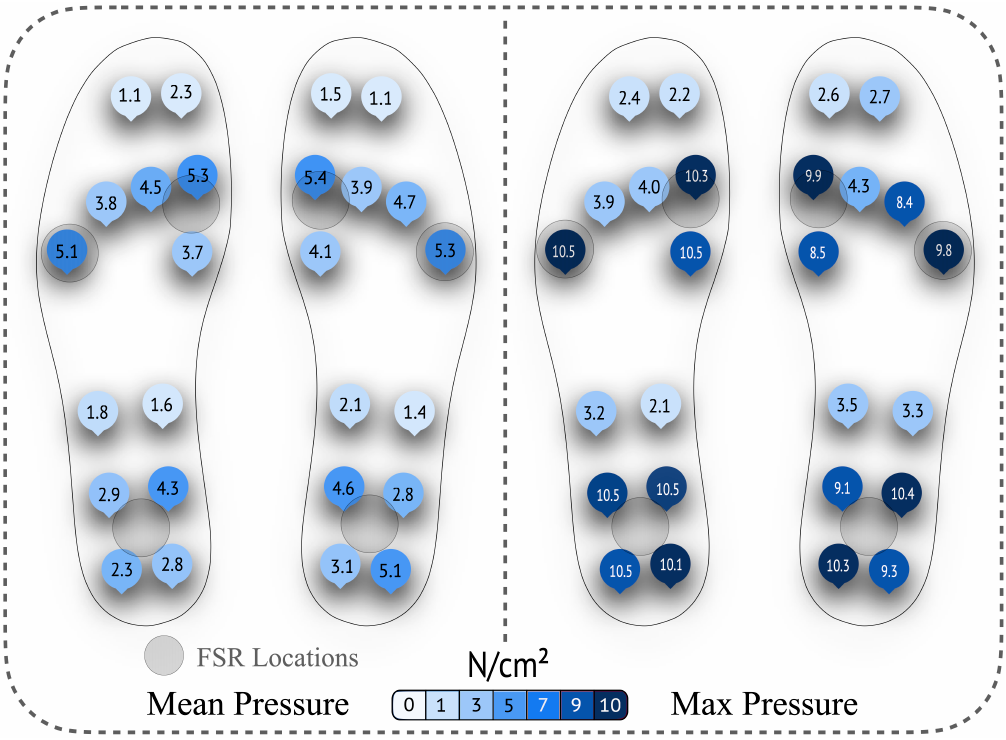}  \\
      (a) & (b)
    \end{tabular}
    \caption{Preliminary tests for informing system design . 
    a) Comparison of GRF components for exoskeleton assisted walking. Multiple crutch strike force data were collected by the force plate and averaged over normalized gait cycles. The mean of each component is plotted along with the standard deviation is illustrated in the shaded region. b) Pressure distribution during exoskeleton-assisted walking. Plantar pressures for 30 gait cycles were recorded using Moticon insoles featuring 14 sensors distributed across the foot. The average mean (left) and maximum (right) pressures for each sensor are denoted.
    }
    \label{fig_preliminaryexperiments}
\end{figure*}

\section{System Description}
The modular device consists of two individual sensor systems (Figure \ref{fig_systemoverview}) that can either work together or separately in order to collect biomechanical data from the exoskeleton. The first part consists of two instrumented crutches able to measure ground reaction forces (GRFs) using load cells, as well as their acceleration and orientation via IMU units. The second part is a pair of 3D-printed insoles which are sensorized with three FSRs each in order to primarily measure the anteroposterior centre of pressure (CoP) of the foot, as well as acceleration and orientation of the foot via an IMU unit. For more information on system design and reproduction manual, please see the supplementary materials.

\subsection{Crutches Design}
Similar to previous works in literature \cite{Gil2021DevelopmentSclerosis}, a pair of standard forearm crutches were cut just above the bottom tip and reinforced with aluminum inner cylinders on both sides, spanning 5 centimeters long, and featuring threaded inputs. The sensor unit is housed in a 3D-printed case and attached with printed clamps on the shaft of the crutch, just under the handle, which minimally shifts the inertia properties of the crutch. Additionally, a flexible elastic 3D-printed hemispherical sleeve was attached at the bottom tips of the crutches to minimize bending moments about the center of the load cell (Figure \ref{fig_preliminaryexperiments}b) and help maintain its integrity.

A load cell LCM200 (Miniature threaded in-line, 250 lb, FUTEK, California, USA) was fitted between the two parts, forming a strong connection through the central axis of the crutch (Figure \ref{fig_systemoverview}b). To avoid high energy requirements and bulky components, an HX711 amplifier (Avia Semiconductor, Xiamen, China) was used to read the load cell’s output and relay the data via its 24-bit analog-to-digital converter to the ESP32 unit. The BNO055 IMU (Bosch, Reutlingen, Germany) was fitted to collect acceleration and orientation data, while the Adafruit ESP32-S3 Feather MCU (Adafruit, New York, USA) was employed for collecting sensor data, featuring a BNO055 IMU, an RGB-LED for visually signaling the state of the peripherals and battery level, and a button for direct physical interaction. A rechargeable 3000 mAh lithium polymer battery powers each sensor board.

To maintain the simplicity, low weight and volume, and cost-effectiveness of the crutches assembly, unlike previous studies where strain gauges were integrated along with bulky braces on the sides of the crutch tip \cite{Sardini2015WirelessMonitoring, Chen2018SmartWalking}, a uni-axial load cell was chosen based on preliminary experiments within this study. In these tests, the crutch GRF contributions in the three axes were examined during exoskeleton-assisted gait for three participants, comparing the force vector components from a force plate by translating them into the crutches’ reference systems. The transverse components were deemed negligible when compared to the central axis of the crutch, ranging from 12.6 ± 3.74 N for mediolateral forces to 24.2 ± 6.78 N for anteroposterior forces, averaging less than 7\% of the vertical forces, as shown in Figure \ref{fig_preliminaryexperiments}a, and agreeing with values previously reported in the literature \cite{Seylan2018EstimationCrutches}.

By exploiting the orientation values from the crutches’ IMUs, we can decompose the resultant GRF into its components and further investigate the anteroposterior and mediolateral components, shedding light on propulsive/braking and balancing forces, respectively. IMU data, including orientation and angular accelerations, and load cell data are serialized to a JSON format. These sensor units conclude the crutches’ peripherals, integrated into the wireless protocol described in subsection C.

\subsection{Insoles Design}
The design of our 3D-printed insoles is primarily driven by considerations of cost-effectiveness, comfort, and ease of application, such as facilitating sensor interchangeability among different insole sizes—small, medium, and large (Figure \ref{fig_systemoverview}b). Our main objective is to investigate the anteroposterior centre of pressure (CoP), thus designs utilizing over 20 sensors to cover the entire foot \cite{Khandakar2022DesignTemperature, Gonzalez2015AnInsoles, Senanayake2010ComputationalGait, Ding2018Proportion-basedInsole} were deemed excessive. Instead, our design adopts a minimalist approach, strategically placing a reduced number of sensors at critical foot locations. As already shown in previous works \cite{Kong2008SmoothPatterns}, gait phases can be detected and categorised with using as low as four FSR sensors, but based on preliminary tests we carried out using the Moticon Insoles (Munich, Germany), we concluded that three sensors would satisfy the requirements of our applications.

A participant used the insoles while walking with the exoskeleton, completing a total of 30 gait cycles. The results were analyzed, showing the highest centres of pressure in the middle of the heel, first, and fifth metatarsal (Figure \ref{fig_preliminaryexperiments}b). In pursuit of simplicity, we positioned three sensors at each of these key locations. This placement aligns with the physical design constraints of lower-limb exoskeletons (LLEs), which typically feature a robust plate at the bottom of the footwear, effectively transforming the foot into a single rigid body driven by ankle flexion. This design assumption supports the division of the foot into two primary contact areas—the heel and the metatarsals—and facilitates a simplified formulation of gait phase estimation outcomes.

The flexible insoles were 3D-printed in small, medium, and large sizes using TPU-90 flexible filament to ensure a solid yet comfortable surface. The insoles feature three gaps each, where key-shaped inserts containing the FSR sensors can be fitted. The bottoms of the insoles have guided ribbon cables that connect to the inserted sensors. This modular design allows for low-cost and efficient sensor interchangeability between different insole sizes. Additionally, if any sensor failure occurs, replacing the sensors is highly efficient.

A 3D-printed clamp mechanism allows for the attachment of the sensor cases to the shoe of the exoskeleton. The analog FSR data are converted by the 12-bit ADC onboard the ESP32-S3 Feather, which also takes in IMU data from the BNO055 included in the sensor unit. Angular accelerations and orientations from IMUs, along with FSR readings, are serialized to a JSON format. These sensor units complete the insole peripherals, integrated into the wireless protocol described in the following subsection. 

While maintaining a lower design complexity and cost than previously developed solutions, our devices offer significant advantages in comfort, modularity, and ease of reproduction. The insoles are not only more comfortable but also modular, allowing for easy sensor interchangeability. The crutches are designed to be lightweight and less bulky, enhancing user convenience. We propose a unified sensor system that can operate together or independently, delivering high-quality results in a more streamlined and cost-effective manner.

\subsection{Central Unit and Data Handling}
The central unit comprises of a Raspberry PI-4 Model-B as to ensure seamless communication and computations between the four sensors. It is powered by a battery pack which is attached on a 3D-printed mounting interface, able to fit on the back of most common LLEs. Wireless communication is established via a Bluetooth Low Energy (BLE) interface facilitated by Bluetooth 4.0, enabling high-speed energy-efficient data transmission. The system allows for both BLE and Bluetooth classic protocols, for higher throughput in data transmission. This enables the peripherals to transmit packets of serialized JSON data to the central, which are then deserialized and written in a tabular CSV file, as well as being fed into a fuzzy logic algorithm. The central system can work we both sensors at the same time, or either one of them independently of the other, based on the desired application. One RGB LED is connected to the Raspberry Pi in order to signal key events such as sensor connection, start and end of recordings, and power status amongst others. 

The raw data collected and stored in the CSV file include the three-dimensional angular accelerations and orientations of the four IMUs, the two amplified load cell values from the crutches, and the six FSR readings from the two insoles, at each timestamp. The frequency and sampling rate of data collection can be easily adjusted as to enable for higher throughput and speed or a bigger transmission distance when the central unit is not on-board the exoskeleton or person. Data transmission can be maintained for up to 30 meters with no interruption, at a maximum frequency of 130 Hz. An Android App facilitates the control of the system via the Bluetooth network. By choosing through a number of inputs, the user can select to calibrate the system, start and end recordings, browse through previous recordings, select which sensors to include in the recording and additionally send triggers to external devices such as camera systems, which can be easily integrated by an accessory module we have created. For more information on handling additional devices please see our Instruction Manual included in the supplementary materials. 

\begin{figure}[th]
    \centering
    \includegraphics[width=1\linewidth]{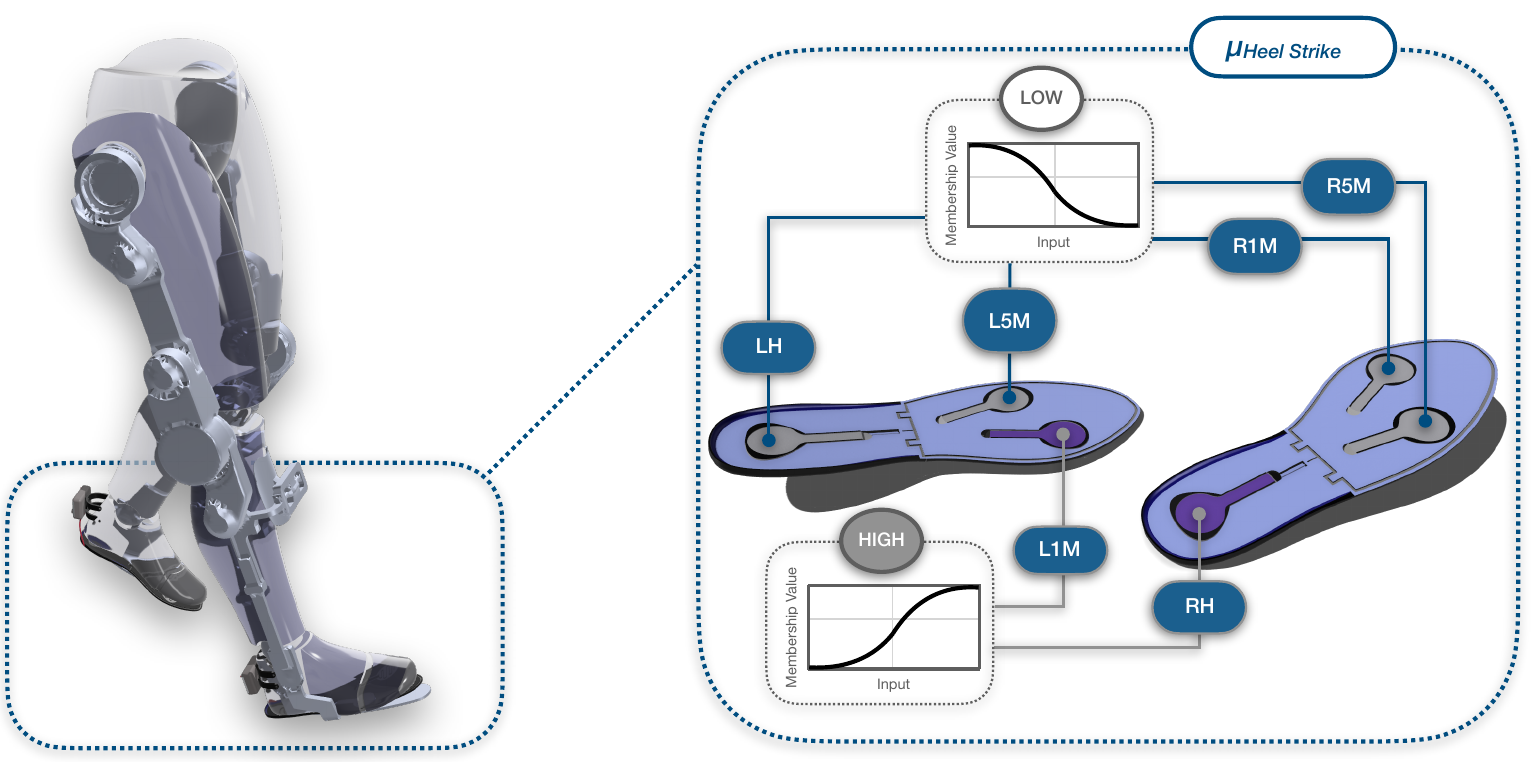} 
    \caption{Example of heel strike through fuzzy logic rule-based approach. During exoskeleton gait, the gait phase is decided through a set of rules based on the membership grades assigned to the FSRs of both insoles using membership functions. Linguistic variables and logical expressions are used in order to decide the outcome, or gait phase. 
    }
    \label{fig_fuzzylogic}
\end{figure}

\subsection{Fuzzy Logic}
Fuzzy logic was employed to process gait values collected from the insoles, aiming to effectively estimate the gait phase. Commonly found works \cite{Chen2019DevelopmentAssistance, Ding2018Proportion-basedInsole}, utilize membership functions expressed as sigmoid functions that evaluate the absolute values of FSR readings,
 \begin{equation}
  \mu(F_i) = \frac{\mathrm{1} }{\mathrm{1} + e^{-s*(F_i - F_{0})} }
  \label{eqn01}
\end{equation}
with $F_i$ being the FSR readings and $\mu(F_i)$ representing the membership grade as result of an FSR reading being evaluated by the membership function.
The parameter $s$ determines the slope of the sigmoid curve and $F_0$ defines the threshold in terms of the absolute value, meaning the point at which $F_1$ yields a membership grade of 0.5.
An option is to utilize half of the maximum possible FSR reading for $F_0$.
The greater $s$, the greater the change in $\mu(F_i)$ with small changes in $F_i$.

During testing with the fuzzy logic algorithm, it was observed that the FSR readings often did not achieve their theoretical maximum values, even under loads exceeding their specified ratings. 
Consequently, this approach of formulating membership functions may lead to less accurate results, that hinder the accuracy of phase estimation.
To mitigate the effects of saturation in FSRs, the readings should be normalised. Other works introduced the normalization of FSR readings for primarily taking into account different walking styles and weight distributions \cite{Chen2019DevelopmentAssistance, Ding2018Proportion-basedInsole}, by scaling down each FSR reading with the sum of all readings.

\begin{equation}
  F_{i}^* = \frac{F_{i}}{\Sigma_{i=1}^nF{i}}
  \label{eqn02}
\end{equation}

Given these circumstances, following membership function is used

\begin{equation}
    \mu(F_i) = \frac{\mathrm{1} }{\mathrm{1} + e^{-0.15*(F_i - 0.45)} }
    \label{eqn03}
\end{equation}

Following the work of Kong and Tomizuka \cite{Kong2008SmoothPatterns}, the linguistic variables 'High' and 'Low', expressed through the aforementioned membership function, are utilized for the fuzzy representation of FSR readings.

\begin{equation}
  \begin{aligned}
    \text{High} &= \mu(F_i) \\
    \text{Low} &= 1 - \mu(F_i)
  \end{aligned}
  \label{eqn04}
\end{equation}

These linguistic variables are used to construct the fuzzy rules that are based on fundamental logical expressions in the form of 

\begin{equation}
  \begin{aligned}
    &\text{IF } F^*_{\text{LH}} \text{ IS LOW AND IF } F^*_{\text{L5M}} \text{ IS HIGH AND} \\
    &\text{IF } F^*_{\text{L1M}} \text{ IS HIGH AND IF } F^*_{\text{RH}} \text{ IS HIGH AND} \\
    &\text{IF } F^*_{\text{R5M}} \text{ IS LOW AND IF } F^*_{\text{R1M}} \text{ IS LOW} \\
    &\text{THEN } G \text{ IS HEELSTRIKE } \\
  \end{aligned}
  \label{eqn03:05}
\end{equation}

\noindent
where $F^*$ designated with the respective subscript represents the scaled-down FSR reading of each sensor placement, while G represents the gait phase. The abbreviations $LH, L1M, L5M, RH, R1M, R5M$ stand for the heel, first metatarsal and fifth metatarsal of the left and right foot respectively. 
Conditional statements, such as $\text{IF } F^*_{\text{LEFT HEEL}} \text{ IS HIGH}$ determine the membership grade of $F^*$ to a particular fuzzy set or linguistic variable, which, in this instance, is $\text{HIGH}$.

Chaining several of these expressions with the $\text{AND}$ operator yields the minimum membership grade, representing the membership grade of G to a particular gait phase. This is expressed by conclusive statements like $\text{THEN } G \text{ IS HEELSTRIKE }$, illsutrated in Figure \ref{fig_fuzzylogic} where the two 'high' sensors are in a darker colour and the rest 'low' sensors in a lighter shade, during a heel strike phase.

\begin{table}[!b]
    \caption{Fuzzy logic rules and gait phase outcomes based on FSR insole membership grades. Outcomes based on right insole.}
    \label{tab_fuzzylogic}
    \centering
    \renewcommand{\arraystretch}{1.25} % Increase space between rows
    \begin{tabularx}{\columnwidth}{c|*{6}{>{\centering\arraybackslash}X}|c}
        \hline\hline
        & \multicolumn{6}{c|}{FSR Grades ($F^*$)} & \multirow{2}{*}{Outcome} \\
        & LH & L5M & L1M & RH & R5M & R1M & \\
        \hline
        & Low   & High & High & High & Low  & High & Heel Strike \\
        \multirow{5}{*}{\rotatebox{90}{Value}} & Low   & Low  & High & High & High & Low  & Loading Response \\
        & Low   & Low  & Low  & High & High & High & Mid-stance \\
        & High  & Low  & Low  & Low  & High & High & Terminal Stance \\
        & High  & Low  & Low  & Low  & Low  & High & Pre-swing \\
        & High  & High & High & Low  & High & Low  & Initial Swing \\
        & High  & High & High & Low  & Low  & Low  & Mid-Swing \\
        & Low   & High & High & Low  & Low  & Low  & Terminal Swing \\
        \hline
    \end{tabularx}
\end{table}

Table \ref{tab_fuzzylogic} presents all rules for gait phase estimation in respect to the right foot. To obtain the rules for the left side, the conditional statements in each rule need to be inversed. 
The presence of red circles signifies the condition $F^* \text{ IS HIGH }$ for a FSR placement. Once all rules are evaluated, each of the eight gait phases is assigned a membership grade.

\begin{table*}[!b]
    \caption{Pearson Correlation Coefficients and RMSE}
    \label{tab_resultsSTATS}
    \centering
    \setlength{\tabcolsep}{16pt} % Increase column separation
    \begin{tabular}{ccccccc}
        \hline\hline % Horizontal line at the top
        & \multicolumn{2}{c}{Anteroposterior Centre of Pressure} & \multicolumn{2}{c}{Crutch Ground Reaction Forces} & \multicolumn{2}{c}{Heel Strike Gait Detection} \\
        & Pearson Corr. & RMSE (mm) & Pearson Corr. & RMSE (N) & Pearson Corr. & RMSE (s) \\ 
        \hline % Horizontal line below headers
        Min & 0.873 & 14.4 & 0.921 & 10.5  & 0.997 & 0.0195  \\ 
        Max & 0.948 & 19.2 & 0.967 & 17.9 & 0.999 & 0.0362 \\ 
        Mean & 0.907 ± 0.038 & 17.2 ± 2.49 & 0.945 ± 0.023 & 15.3 ± 4.21 & 0.998 ± 0.001 & 0.0291 ± 0.0084 \\ 
        \hline % Horizontal line at the bottom
    \end{tabular}
\end{table*}

\section{Validation Experiments}
Three different experiments took place in order to validate hardware and software capabilities of the system, against gold standard marker motion capture (Qualisys, Gothenburg, Sweden) and force plate (Kistler Group, Winterhur, Switzerland) technologies. The first two experiments comprised of standalone equipment testing, while for the third experiment three able-boded participants (1 female and 2 males, age 36 ± 7 years, weight 68.7 ± 12.1 kg, height 1.67 ± 0.08 m) were recruited. The participants performed a set of six walking bouts each, with the TWIN lower-limb exoskeleton \cite{Laffranchi2021}. All participants signed informed consent forms prior to the experiments, the procedures of which were in accordance with the Declaration of Helsinki and approved by the Ethical Committee of Heidelberg University (resolution S-313/2020). 

All post-processing computations, data and statistical analyses were conducted using MATLAB (MathWorks Inc., Natick, Massachusetts, USA). For synchronization of data recordings, a trigger sub-system was created using an ESP32 Firebeetle which was connected to the Qualisys system, incorporating the force plates, and sent a trigger signal through at the start of every recording.

\subsection{Anteroposterior Centre of Pressure}
The center of pressure is a critical metric for assessing balance and understanding the intentions—whether propulsive, braking, or stationary—of an exoskeleton user. For estimating gait phases and detecting user motion-intention, the anteroposterior direction (the direction of travel) plays the most critical role. Consequently, the anteroposterior CoP, measured along the length of the foot, was prioritized in our analysis of the system's capabilities. To capture the entire area of the feet, a person placed one foot on a force plate and performed circular clockwise motions, shifting their center of mass to record comprehensive CoP data. This circular motion was performed 18 times across three separate experiments.
The CoP was then calculated from the FSR data of the insoles by computing the weighted averages for mediolateral and anteroposterior coordinates using the locations of the FSRs,

\begin{equation}
    y_{\text{CoP}} = \frac{(y_H \cdot F_H) + (y_M1 \cdot F_M1) + (y_M5 \cdot F_M5)}{F_{\text{total}}}
    \label{eqn06}
\end{equation}

\noindent 
where \( y_H, y_M1, y_M5 \) are the anteroposterior coordinates of the heel, first metatarsal, and fifth metatarsal sensors, respectively; \( F_H, F_M1, F_M5 \) are the FSR values measured at these points; and \( F_{\text{total}} \) is the sum of these values. Subsequently, the force plate CoP was transformed using the markers on the shoe from the global laboratory reference system to the insole reference system, as to enable comparison between the two values. 

\subsection{Crutches Ground Reaction Forces}
A crucial yet often overlooked metric in the assessment of exoskeleton-assisted gait is the contribution of the upper body. To address this gap, our instrumented crutches are designed to provide detailed insights into how users utilize forearm crutches to support their gait. In one of the experiments, a person repeated a total of 18 circular movements over three experiments while applying forces to the crutch and maintaining it in contact with a force plate. The resulting force vector, captured by the force plate, was then translated into the crutch's reference system using markers placed on the crutch. The data from the load cells were subsequently compared to the translated vertical component of the force plate, which aligned with the crutch's central axis. 

\subsection{Heel Strike Gait Detection}
To assess the capability of our sensors to accurately estimate gait phases, we focused on a fundamental variable: the heel strike. Three participants, each equipped with the system and markers, performed 18 trials across the laboratory using the exoskeleton. By analyzing the data from the heel FSRs, we identified the time frames of local maxima, which were then compared to those captured by motion capture cameras. Heel strikes from markers were defined by calculating the maximum and minimum distances between toe and sternum, and heel and sternum markers, as outlined in our prior research \cite{Marinou2022TowardsFamiliarization}. Toe-off events were also analyzed to provide further comparative data and enhance the evaluation of our system's robustness.

\subsection{Data and Statistical Analysis}
Data were collected from our custom system, motion capture cameras, and force plates at sampling rates of 130 Hz, 150 Hz, and 1500 Hz, respectively. To ensure clarity and remove noise, a Butterworth filter with respective orders and cut-off frequencies was applied to each data stream separately. For comparative analysis, the data from higher sampling rates were downsampled to match the lowest frequency data, thus normalizing the datasets. This approach to data treatment preserved the integrity and essential features of the raw data, ensuring an accurate representation for further analysis. 

For our statistical analysis, a Shapiro-Wilk test at a significance level of $\alpha = 0.05$ revealed a non-normally distributed nature for our data sets. This led to all of our comparisons between paired data sets of biomechanical metrics collected using two different systems, to be made using the non-parametric Wilcoxon Signed-Rank test, where $ p < 0.05$. Hence, our data is reported as median and interquartile range (IQR) and based on the significance of the results, we report the median and IQR values for the differences between different collection methods, and the minimum, maximum, mean and standard deviations (STD) of root-mean-square error (RMSE) and the Pearson correlation coefficients between the data sets. Results are reported for the left side insole and crutch, as the Wilcoxon Signed-Rank test between the two sides did not show a significant difference. 

\begin{figure*}[htbp]
    \centering
    \begin{minipage}{.5\linewidth}
        \centering
        \includegraphics[width=\linewidth]{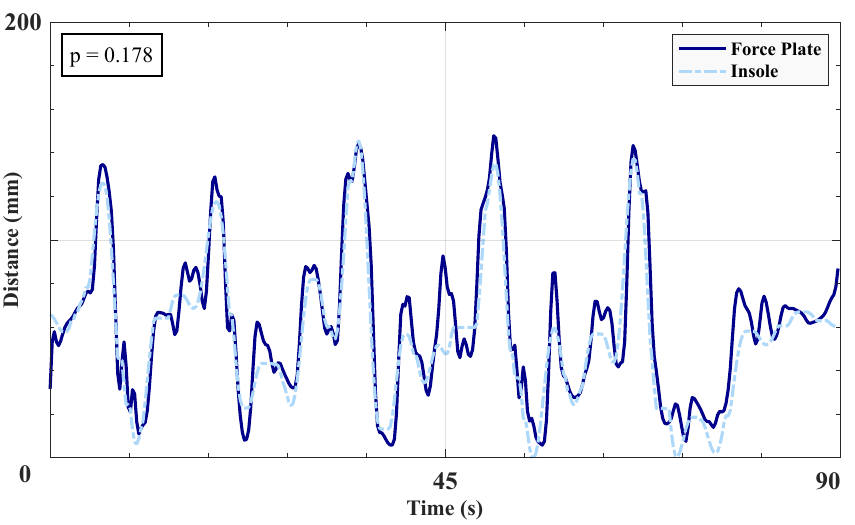}
        \caption*{(a)}
    \end{minipage}%
    \begin{minipage}{.5\linewidth}
        \centering
        \includegraphics[width=\linewidth]{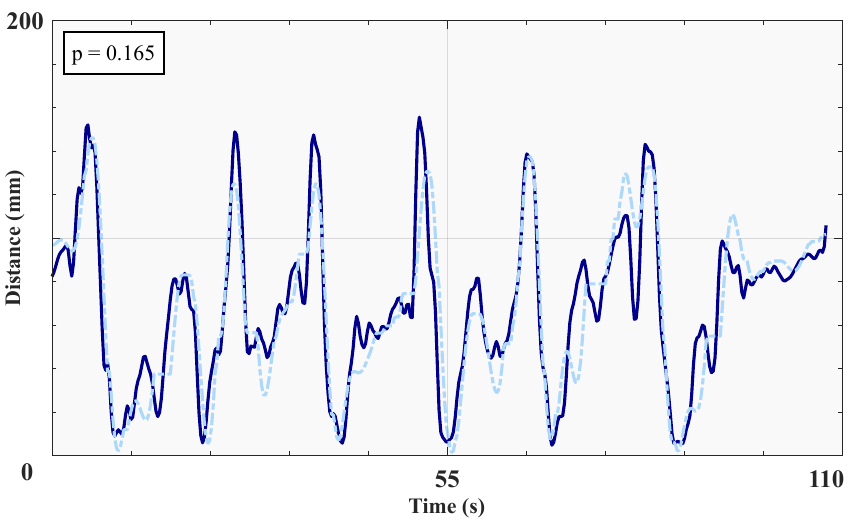}
        \caption*{(b)}
    \end{minipage}
    
    \begin{minipage}{.5\linewidth}
        \centering
        \includegraphics[width=\linewidth]{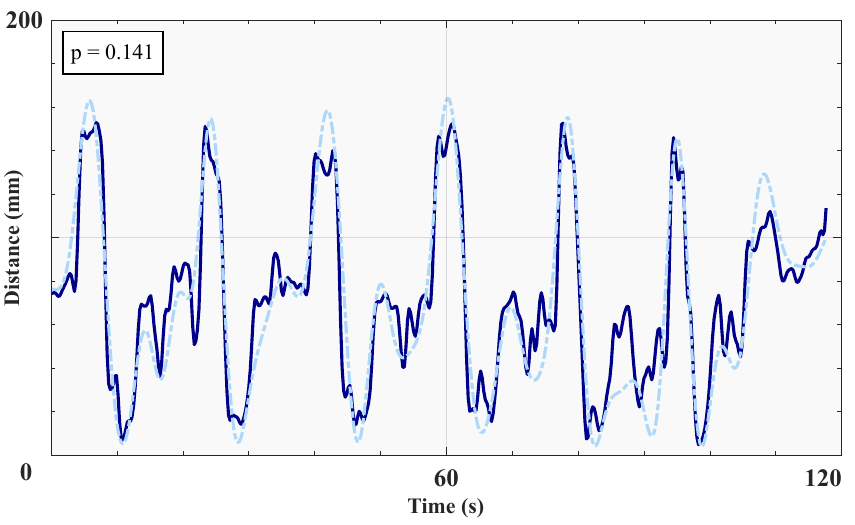}
        \caption*{(c)}
    \end{minipage}%
    \begin{minipage}{.5\linewidth}
        \centering
        \includegraphics[width=\linewidth]{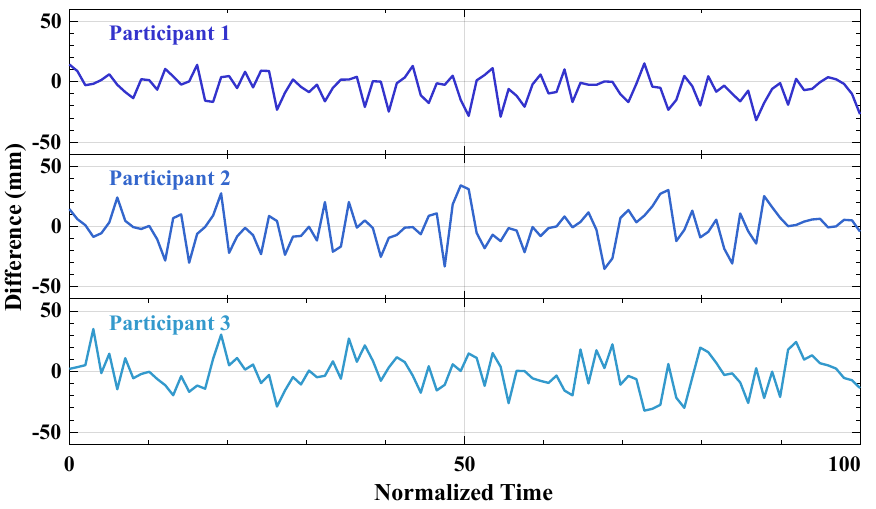}
        \caption*{(d)}
    \end{minipage}
    
    \caption{Anteroposterior centre of pressure comparison between insole and force plate. (a) - (c) Three time series of the anteroposterior CoP as recorded by force plate and insole, from each participant in order, performing circular motions about their foot. $p$ values are provided for each trial. (d) Actual values of the anterposterior CoP differences between the force plate and the insole, for all three trials separately.}
    \label{fig_resultsCoP} 
\end{figure*}

\begin{figure*}[htbp]
    \centering
    \begin{minipage}{.5\linewidth}
        \centering
        \includegraphics[width=\linewidth]{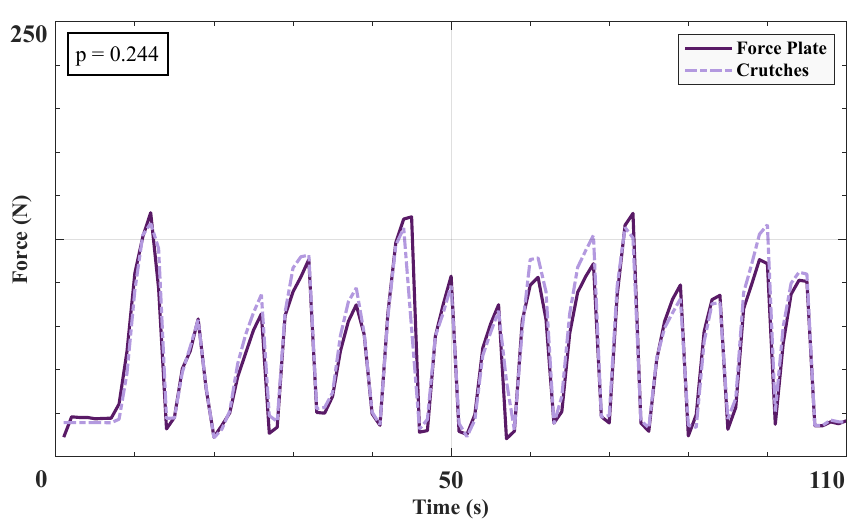}
        \caption*{(a)}
    \end{minipage}%
    \begin{minipage}{.5\linewidth}
        \centering
        \includegraphics[width=\linewidth]{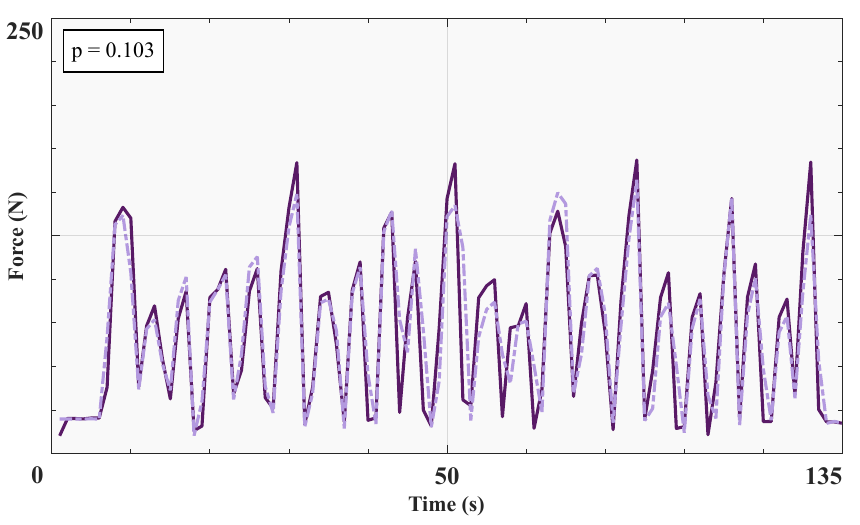}
        \caption*{(b)}
    \end{minipage}
    
    \begin{minipage}{.5\linewidth}
        \centering
        \includegraphics[width=\linewidth]{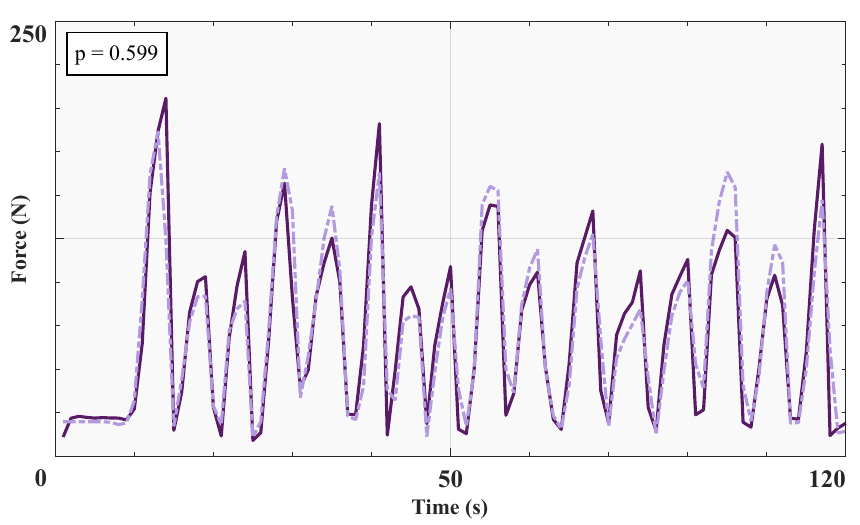}
        \caption*{(c)}
    \end{minipage}%
    \begin{minipage}{.5\linewidth}
        \centering
        \includegraphics[width=\linewidth]{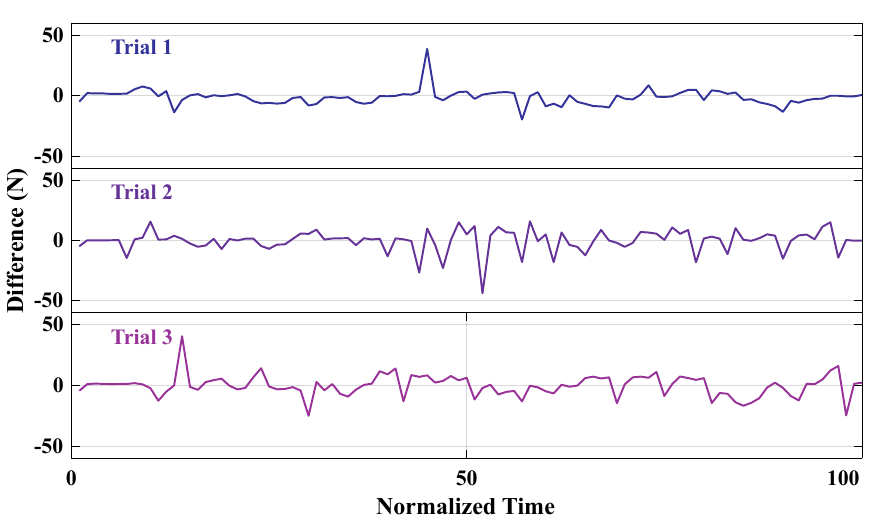}
        \caption*{(d)}
    \end{minipage}
    
    \caption{Ground reaction forces comparison between crutch and force plate. (a) - (c) Three time series of the GRFs as recorded by force plate and insole, from each participant in order, performing circular motions about their foot. $p$ values are provided for each trial. (d) Median and IQR values of the GRFs differences between the force plate and the insole, over all three trials.}
    \label{fig_resultsGRFs} 
\end{figure*}

\begin{figure*}[htbp]
    \centering
    \begin{minipage}{1\linewidth}
        \centering
        \includegraphics[width=\linewidth]{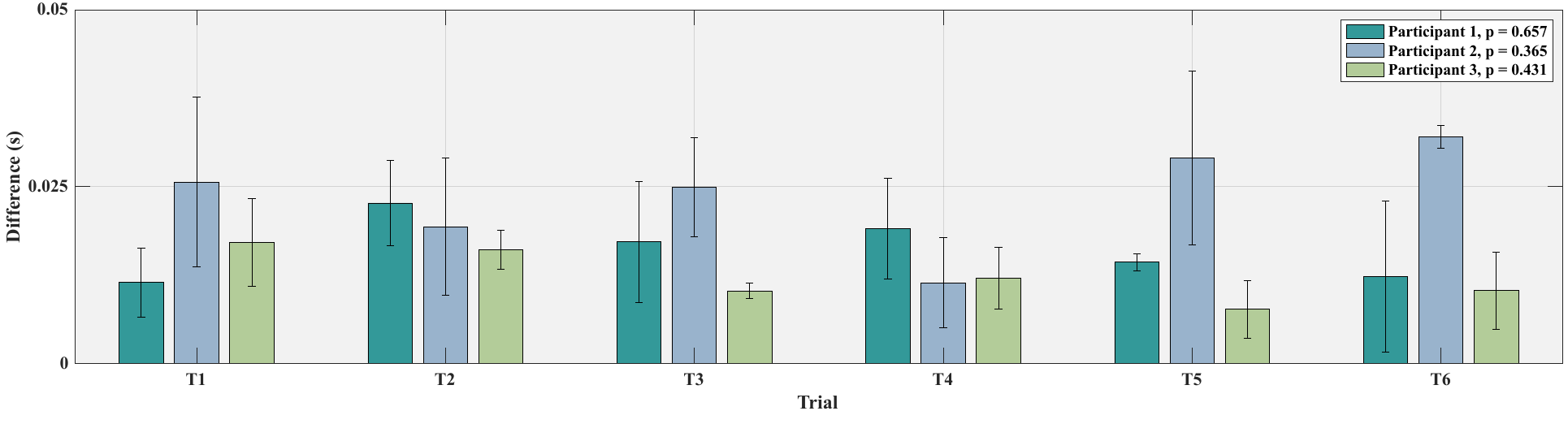}
        \caption*{(a)}
    \end{minipage}%
    
    \begin{minipage}{1\linewidth}
        \centering
        \includegraphics[width=\linewidth]{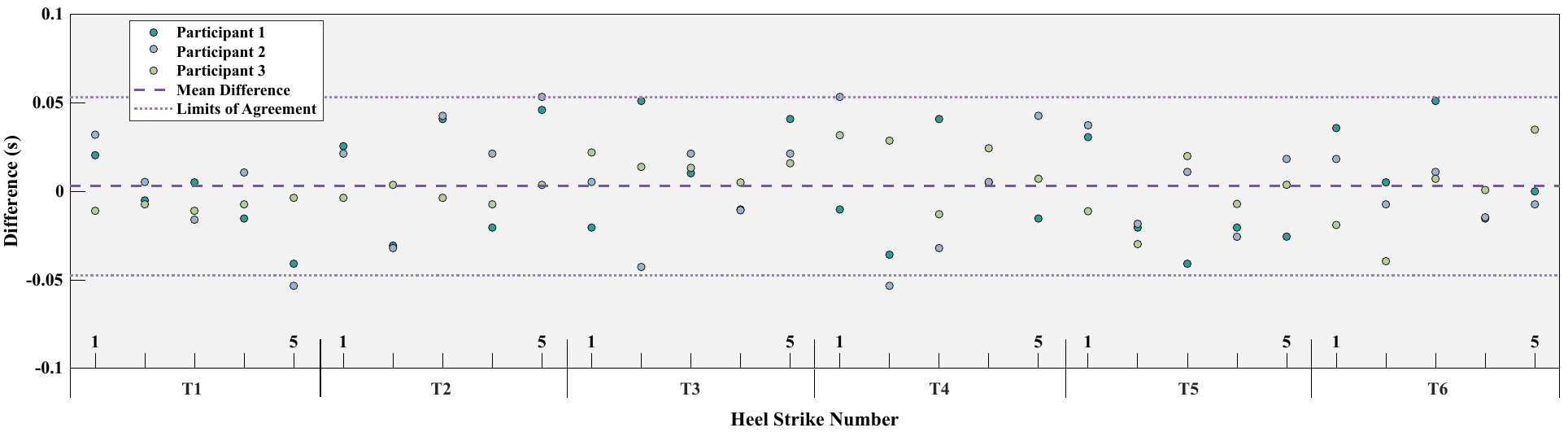}
        \caption*{(b)}
    \end{minipage}
    \caption{Heel strike comparison between insoles and markers. (a) Bar plots for six trials reporting the median and interquartile ranges of the differences in heel strikes for each of the three participants. $p$ values reported in the legend for each participant. (b) Difference of markers and insole heel strikes plotted against, for each participant heel strike over six trials, along with the median difference and limits of agreement.}
    \label{fig_resultsHSs} 
\end{figure*}

\section{Results and Discussion}
Lower-limb exoskeletons have advanced wearable technology by significantly restoring mobility and rehabilitating impaired gait for individuals with movement disorders \cite{Terrazas-Rodas2022Lower-LimbReview}. However, their efficient integration into daily life requires thorough biomechanical assessment, typically reliant on costly lab-based equipment that fails to capture critical biomechanical parameters in practical settings. Moreover, achieving transparent control remains challenging, with current solutions often depending on expensive sensors or complex machine learning techniques that may lack adequate user feedback and comfort.

To address these challenges in a cost-effective and user-centered manner, we developed a modular sensor-based system for typical LLEs, capable of recording key biomechanical metrics for assessment and motion intention-based control. The system includes forearm crutches with load cells and flexible 3D-printed insoles with pressure sensors, both integrated with IMUs. Data is processed through a fuzzy logic algorithm for efficient and accurate gait phase estimation. Additionally, we provide full open-source access to our hardware and software designs, along with instruction manuals. This approach advances real-life applications of LLEs, and to the authors' knowledge, these technologies have not been combined before, particularly in such a low-cost, simple, and effective manner.

To validate our system and the accuracy of the data recorded from the sensors, we performed a set of validation experiments against force plate and motion capture systems, based on three different metrics. Figures \ref{fig_resultsCoP} to \ref{fig_resultsHSs} visualize the comparative performance of our system against these gold-standard approaches, and Table \ref{tab_resultsSTATS} summarizes key statistical outcomes extracted from these three metrics.

\subsection{Anteroposterior Centre of Pressure}
The plantar CoP is a crucial metric for both assessing balance and informing the control of LLEs, with the AP component indicating whether the user intends to move forward or remain stationary. The AP coordinates of the CoP as recorded from the force plate and the insoles are shown in Figure \ref{fig_resultsCoP}, parts A to C, for each of the three trials separately, as well as the actual difference values in part D, for each trial separately. Table \ref{tab_resultsSTATS} shows that the mean RMSE of the difference between the left insole and the force plate measurements over all three trials is 17.2 ± 2.50 mm. This indicates a high degree of accuracy, suggesting that the insole measurements closely match those of the force plate.The p-values for the first, second, and third trials are $0.178$, $0.165$, and $0.141$, respectively, indicating no statistically significant difference between the insole and force plate measurements. A high Pearson coefficient of $0.907 \pm 0.038$ indicates a close correlation between the two measurement technologies.

Based on these results, the anteroposterior centre of pressure measured by our insoles demonstrates a high degree of accuracy when compared to force plate data, comparable to previous research \cite{Jonsson2019FootSystem} (Pearson correlation coefficients $0.84 - 0.90$). For analyzing gait intentions and determining whether specific sensor membership grades vary from high to low, our system is more than adequate, as shown by the close matches of high and low peaks in Figure \ref{fig_resultsCoP}. Additionally, the high level of accuracy provided by the insoles offers a reliable biomechanical metric for assessing balance in assisted gait, achieved through a minimal array of three sensors.

\subsection{Crutches Ground Reaction Forces}
Upper body contributions in exoskeleton-assisted gait are often overlooked, thus the effort exerted by the user cannot truly be quantified as a whole. For long-term use of assistive exoskeletons, upper body efforts need to be evaluated and minimized to achieve an energy-efficient integration of exoskeletons into daily life \cite{Ugurlu2014LowerEffort}. Forearm crutches with GRFs provide important insight into how much upper body effort users exert when using LLEs. Figure \ref{fig_resultsGRFs} reports the GRFs recorded from the left crutch for each of the three trials separately (A - B) and also for the median and IQR of the differences between the two collection systems over the three trials. Values for the Pearson coefficients ($0.945 \pm 0.023$) and RMSEs ($15.3 \pm 4.21$) are reported in Table \ref{tab_resultsSTATS}. The mean RMSE between the force plate and crutch is $15.3 \pm 4.21$ N over the three trials, and the $p$ values for each trial are $0.244$, $0.103$, and $0.599$.

The GRFs recorded from the instrumented crutches achieved a very high level of accuracy when compared to values obtained from force plates, as shown both in Figure \ref{fig_resultsGRFs} and Table \ref{tab_resultsSTATS}. The deviation of the crutches was minimal, and the percentage mean RMSE of the ground-truth GRF ($4.84\% $) is comparable to similar instrumented crutches studies using more comprehensive sensor arrays \cite{Seylan2018EstimationCrutches,Sardini2015WirelessMonitoring} ($2\% - 5.4\%$ mean RMSE of total crutch GRF). In conclusion, our instrumented crutches offer a cost-effective and reliable method for accurately investigating upper body contributions via GRFs in LLE-assisted gait.

\subsection{Heel Strike Gait Detection}
Whether assessing key biomechanical metrics such as gait speed and stride duration, or accurately segmenting gait cycles from beginning to end, the heel strike is arguably the most important phase indicator \cite{Roberts2017BiomechanicalGait}. The heel strike timestamps detected from markers on the left shoe and the FSRs of the left insole are compared in Figure \ref{fig_resultsHSs}. Part A reports the median and IQR values for each participant and each of the six trials separately, whereas part B visualizes the errors for each individual heel strike over the three participants, as compared to the heel strikes calculated from marker data. 

The mean RMSE between the markers and the insole over the six trials for all three participants is $0.0291 \pm 0.0084$ s, as reported in Table \ref{tab_resultsSTATS}, highlighting the insole’s high precision and robustness. This corresponds to an mean absolute error of 28.1 milliseconds which indicates high precision and performance of the system, especially in the context of exoskeleton gait. When compared to marker-derived timestamps of heel strikes, our method demonstrated a mean error of 0.844\% ± 0.317\% relative to the respective step durations, where the mean step duration with the exoskeleton was 3.33 ± 1.16 seconds. 

Figure \ref{fig_resultsHSs}b shows close agreement between marker-calculated heel strikes and insole, supported by a very strong Pearson correlation of $0.998 \pm 0.001$ across all participants. The limits of agreement present on the figure correspond to ±1.96 SD (95\% confidence), and the equal spread of data below and above the mean highlight that the errors were symmetrically distributed around the true values, indicating no systematic bias in the method. Comparable to previous studies using insoles to identify heel strikes and segment gait, with a mean absolute error of $0.0168 s$ (\cite{Kim2020GaitInsoles, Crea2014AAnalysis} mean absolute errors $0.01$ to $0.03 s$), our insoles demonstrate high precision, which validates the calculation of gait phase duration, as well as the use of FSR sensors within the fuzzy logic context for a rule-based approach to gait phase estimation.

It is important to note that this study can be improved by testing the system on a larger number of participants to ensure stronger statistical significance of results. Additionally, of utmost importance is the implementation of the proposed control strategies with an LLE and the validation of the system's efficacy in real-time use of exoskeletons. Future studies will focus on implementing these controllers and refining software control strategies based on the preliminary outcomes presented in this study.

This research established a preliminary test of our prototyped system to gather real-time biomechanical data and inform gait analysis and control of exoskeletons. We have demonstrated a very strong correlation of basic biomechanical metrics calculated using our system, compared to gold-standard force plate and motion capture systems' ground truths. Our results are comparable, if not superior, to current alternatives, offering distinct advantages of low production costs and design simplicity. Additionally, all of our work is provided open-source to ensure reproducibility and foster collaborative improvements in future iterations.

\section{System Applications}
Our modular sensor system offers a versatile platform for advancing the functionality and assessment of lower-limb exoskeletons. This section explores the two primary applications of our system: biomechanical evaluation and high-level control. By leveraging a comprehensive array of biomechanical metrics and sophisticated control algorithms, our system does not only offer the potential to enhance the understanding of exoskeleton-user interactions but also significantly contributes to the development of responsive and adaptive assistive technologies.

\subsection{Biomechanical evaluation}
Integrated load cells and IMUs on crutches capture upper body dynamics during gait, offering data on propulsive, braking, and balancing GRFs essential for exoskeleton assessment and improvement. Similarly, insoles equipped with force-sensitive resistors measure the center of pressure (CoP) trajectories, crucial for analyzing balance and stability in real-time across various gait scenarios. Key metrics such as stride duration and gait speed are also derived, enabling comprehensive gait analysis. Orientation and acceleration data from IMUs provide detailed insights into limb positions and movements throughout the gait cycle.

Our system uniquely combines crutches' IMUs and insole sensors to accurately determine distinct gait phases, offering the capability of aligning exoskeleton mechanics with the user’s natural movements and enhancing device responsiveness. A significant advantage of our system is its capability to perform evaluations in non-laboratory environments, broadening the scope for real-world exoskeleton testing and adaptation. This feature is invaluable for the iterative design and refinement of exoskeletons tailored for daily use. Its open and adaptable design ensures it can be customized to meet specific research or clinical demands, fostering innovation and continuous improvement.

\begin{figure}[th]
    \centering
    \includegraphics[width=1\linewidth]{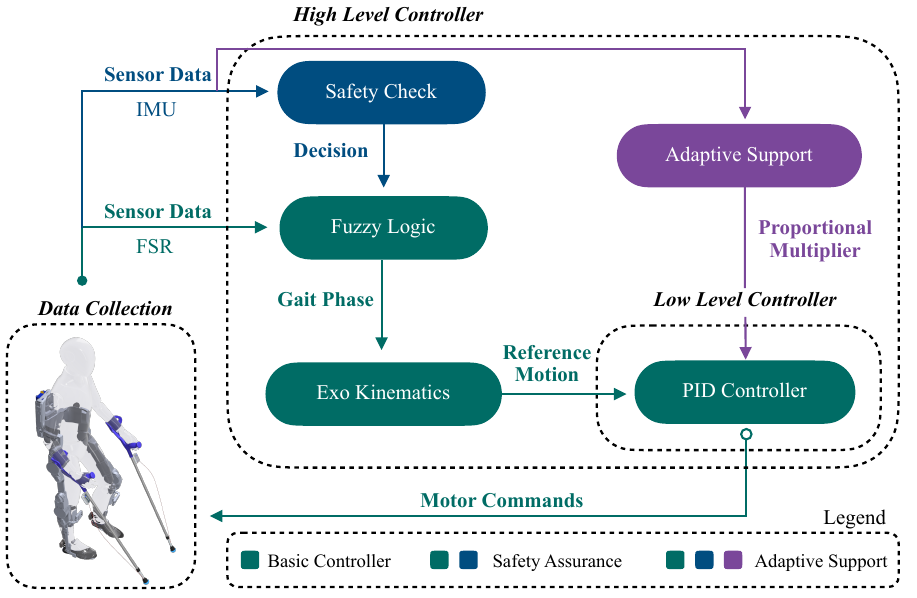} 
    \caption{Proposed controllers example. The green schematic describe the normal processes of the basic controllers, whereas the additional blue controls outline the extra functions incorporated by the safety assurance controller, within the higher level controller of the system. If the decision made from the safety check proves true (or safe) then the basic controller processes take place. Finally, the purple schematics introduce the proportional multiplier directly applied onto the PID controller of the lower level control of the exoskeleton. 
    }
    \label{fig_controllers}
\end{figure}

\subsection{High-level controller}
Real-time control of exoskeletons can be achieved by utilizing the variables extracted from our modular sensor system, via fuzzy logic as to achieve precise motion intention detection and provide adequate assistance to the user. Various control strategies can be employed using different subsystem configurations. In this section we provide proposals for three simple feedback-based control schemes that can be created from the collected data.

\subsubsection{Basic Gait Phase Detection Controller}
A basic control loop utilizes a simple fuzzy logic approach based on FSR voltages to identify distinct gait phases. The controller processes the sensor data to construct membership functions that determine the current phase of the gait cycle. Once a specific gait phase is recognized, the controller sends a signal to the exoskeleton to initiate or adjust movement, ensuring that the device is synchronized with the user’s natural walking pattern. This basic control loop is crucial for smooth operation and enhances the natural feel of the exoskeleton during use.

\subsubsection{Safety Assurance Controller}
Building on the basic gait phase detection, a safety assurance controller incorporates additional safety checks using sensor data to verify the orientation and stability of both crutches and insoles. Before executing any movement commands, the system checks that both crutches are in contact with the ground confirming stable support, both feet have executed forward movements as indicated from IMU accelerations, and the orientation of the crutches and insoles is within predefined safe ranges to prevent falls or improper loading. This safety controller ensures that all conditions are met before allowing the exoskeleton to proceed with movement, significantly reducing the risk of accidents and enhancing user confidence.

\subsubsection{Adaptive Support Controller}
The most advanced control loop is the adaptive support controller, which not only recognizes gait phases and ensures safety but also adjusts the level of support provided by the exoskeleton in real-time. This controller uses GRF data from crutches to modulate the support intensity; higher forces detected via the load cells lead to increased assistance, adapting to the user’s exertion level.
IMU data is used to gauge the inclination of the crutches, which correlates with walking speed intentions. A lower inclination angle could indicate a desire to walk faster, prompting the controller to decrease the response time of support mechanisms.
This proportional control strategy ensures that the exoskeleton’s assistance is both responsive and attuned to the user’s immediate needs, providing a tailored support experience that adjusts dynamically to changes in walking dynamics or user fatigue.

\section{Conclusion}
This study addressed the challenges of cost-effective and user-centered integration of lower-limb exoskeletons (LLEs) into daily life by developing a modular sensor-based system capable of recording key biomechanical metrics for both assessment and motion intention-based control. The system, comprising force-sensing forearm crutches and 3D-printed insoles with integrated IMUs and pressure sensors, utilizes a fuzzy logic algorithm for efficient gait phase estimation. Validation experiments against force plate and motion capture systems demonstrated high accuracy in measuring the anteroposterior center of pressure, ground reaction forces, and heel strike detection. Our results, which showed strong correlations and low RMSE values, confirmed that the system effectively captures critical biomechanical parameters. The open-source availability of our hardware and software designs further promotes innovation and broader application in exoskeleton research. By meeting its objectives, this study significantly advances the practical usability and assessment of LLEs, fostering further development in wearable robotics and assistive technology.

\section*{Acknowledgments}
The authors would like to thank the Carl Zeiss Foundation (Germany) for providing funding through the HeiAge project to carry out this research work.

\section*{Publication Statement}
This work has been submitted to IEEE Transactions in Medical Robotics and Bionics awaiting review. 

\bibliographystyle{IEEEtran}
\bibliography{IEEEabrv,references}{}

\vfill

\end{document}